\pdfoutput=1

\documentclass[11pt]{article}

\usepackage[]{acl}

\usepackage{times}
\usepackage{latexsym}

\usepackage{dcolumn}
\newcolumntype{L}{D{.}{.}{2,3}}

\usepackage[T1]{fontenc}

\usepackage[utf8]{inputenc}

\usepackage{microtype}

\usepackage{graphicx}
\usepackage{array}
\usepackage{subcaption}
\usepackage{latexsym}
\usepackage{multirow}

\usepackage{booktabs}
\usepackage{colortbl}
\usepackage{xcolor}

\definecolor{light-gray}{gray}{0.9}

\usepackage[shortlabels]{enumitem}

\title{Evaluating Factuality in Text Simplification}

\author{Ashwin Devaraj$^1$\ \  
William Sheffield$^{2,4}$\ \ 
Byron C.\ Wallace$^3$\ \ 
Junyi Jessy Li$^4$\\
$^1$ Computer Science, $^2$ Mathematics, $^4$ Linguistics, The University of Texas at Austin\\
$^3$ Khoury College of Computer Sciences, Northeastern University\\
{\small \tt ashwin.devaraj@utexas.edu,}
{\small \tt sheffieldw@utexas.edu}\\
{\small \tt b.wallace@northeastern.edu,}
{\small \tt jessy@utexas.edu}\\
}

\date{}

\begin{document}
\maketitle
\begin{abstract}

Automated \textit{simplification} models aim to make input texts more readable. Such methods have the potential to make complex information accessible to a wider audience, e.g., providing access to recent medical literature which might otherwise be impenetrable for a lay reader. However, such models risk introducing errors into automatically simplified texts, for instance by inserting statements unsupported by the corresponding original text, or by omitting key information.
Providing more readable but inaccurate versions of texts may in many cases be worse than providing no such access at all. The problem of factual accuracy (and the lack thereof) has received heightened attention in the context of summarization models, but the factuality of automatically simplified texts has not been investigated. We introduce a taxonomy of errors that we use to analyze both references drawn from standard simplification datasets and state-of-the-art model outputs. We find that errors often appear in both that are not captured by existing evaluation metrics, motivating a need for research into ensuring the factual accuracy of automated simplification models.

\end{abstract}

\section{Introduction}

Simplification methods aim to make texts more readable without altering their meaning. This may permit information accessibility to a wide range of audiences, e.g., non-native speakers~\cite{yano1994effects}, children~\cite{de2010text}, as well as individuals with aphasia~\cite{carroll1998practical} and dyslexia~\cite{rello2013frequent}. Simplification may also help laypeople digest technical information that would otherwise be impenetrable~\cite{damay2006simtext,devaraj2021paragraph}. 

Recent work has made substantial progress by designing sequence-to-sequence neural models that ``translate'' complex sentences into simplified versions~\cite{sari,alva2020data}. An important but mostly overlooked aspect of automated simplification---especially in the conditional text generation regime---is whether outputs are \emph{faithful} to the inputs that they are simplifying. Consider, for example, automatically simplifying medical texts \cite{devaraj2021paragraph}: Presenting individuals with readable medical information that contains factual errors is probably worse than providing no such access at all.

\begin{table}[t]
    \centering
    \small
    \begin{tabular}{p{0.1cm} p{7cm}}
    \toprule
    (1) & \textbf{[Original]} There was no difference in operating time or perioperative complication rates. \\
    & \textbf{[Model simplified]} However, there was not enough evidence to determine if there was an important difference in operative time or complication rates \textcolor{red}{when compared to conventional surgery.} \\
    \midrule
    (2) & \textbf{[Original]} All studies were associated with methodological limitations.\\
    & \textbf{[Model simplified]} All studies \textcolor{red}{were of poor quality} and had limitations in the way they were conducted. \\
    \midrule
    (3) & \textbf{[Original]} On June 24 1979 (the 750th anniversary of the village), Glinde \textcolor{blue}{received its town charter}.\\
    & \textbf{[Model simplified]} On June 24 1979, the 750th anniversary of the village \textcolor{red}{was renamed}. \\
    \midrule
    (4) & \textbf{[Original]} Others agreed with the federal court; they started \textcolor{blue}{marrying} people in the morning.\\
    & \textbf{[Model simplified]} Others agreed with the federal court; they started \textcolor{red}{trying} in morning.\\
    \midrule
    (5) & \textbf{[Original]} In 2014, Mary Barra became CEO of General Motors, making her the first female CEO of a major automobile company.\\
    & \textbf{[Model simplified]} Also, just one woman leads a major automobile company. \textcolor{red}{\emph{Omitted main subject.}} \\
    \bottomrule
    \end{tabular}
    \vspace{-0.5em}
    \caption{Original texts from the Wiki, news, and medical domains with corresponding outputs from simplification systems. Models introduce factual errors.}
    \label{tab:introexamples}
\vspace{-0.5em}
\end{table}

Recent work has acknowledged factuality and faithfulness as key issues to be addressed in other conditional generation tasks like summarization~\cite{kryscinski2020evaluating,maynez2020faithfulness,pagnoni2021understanding,goyal2021annotating}, yet so far little research has thoroughly studied the kinds of errors that simplification datasets and system outputs exhibit. This work seeks to close this research gap.

Table~\ref{tab:introexamples} shows examples of generated outputs from existing simplification systems, and these clearly illustrate that factuality is an issue. We conduct multi-dimensional analyses based on the edit nature of simplification~\cite{xu2015newsela,dong-etal-2019-editnts} and define a small typology of (potential) factual errors in the context of simplification. \emph{Inserting} information can be useful to define jargon and provide explanatory content, but introducing irrelevant or erroneous content (``hallucinating'') is bad (e.g., examples 1-2 in Table~\ref{tab:introexamples}). \emph{Omitting} information related to the main entity or event could lead to a change in how the text is understood (e.g., example 5 in Table~\ref{tab:introexamples}). Finally, making inappropriate \emph{substitutions} can result in inconsistencies (e.g., examples 3-4 in Table~\ref{tab:introexamples}). Together these dimensions represent the precision, recall, and accuracy of information conveyed in simplified texts.

We collect human ratings of factuality for these aspects on two widely used simplification corpora: Wikilarge~\cite{zhang2017sentence} and Newsela~\cite{xu2015newsela}. Automatically aligned sentences from these two datasets are typically used to train and evaluate supervised simplification systems.
We find that errors occur frequently in the validation and test sets of both datasets, although they are more common in Newsela (Section~\ref{sec:Results_data}).

We then evaluate outputs from several modern simplification models~\cite{zhang2017sentence,dong-etal-2019-editnts,access,controlts}, as well as a fine-tuned T5~\cite{t5} model. Compared to RNN-based models, Transformer-based ones tend to have less severe deletion and substitution errors; however, the pre-trained T5 produced more hallucinations on the more abstractive Newsela dataset. We find that existing quality metrics for simplification such as SARI~\cite{sari} correlate poorly with factuality. Although deletion errors correlate with existing semantic similarity measures, they fail to capture insertion and substitution.

As an initial step towards automatic factuality assessment in simplification, we train  RoBERTa~\cite{roberta}-based classification models using our annotated data, and use synthetically generated data to supplement training. We demonstrate that this is a challenging task.

Our code and data can be found at \href{https://github.com/AshOlogn/Evaluating-Factuality-in-Text-Simplification}{https://github.com/AshOlogn/Evaluating-Factuality-in-Text-Simplification}.

\section{Related Work}

Factuality (and the lack thereof) has been identified as critical in recent work in unsupservised simplification~\cite{laban-etal-2021-keep} and medical simplification~\cite{devaraj2021paragraph}. \citet{guo2018dynamic} incorporated textual entailment into their simplification task via an auxillary loss. They showed that this improved simplifications with respect to standard metrics and human assessments of output fluency, adequacy, and simplicity, but they did not explicitly evaluate the resultant factuality of outputs, which is our focus.

Given the paucity of prior work investigating factuality in the context of automated simplification, the most relevant thread of research to the present effort is work on measuring (and sometimes improving) the factuality in outputs from neural \emph{summarization} systems. \citet{falke2019ranking} proposed using textual entailment predictions as a means to identify errors in generated summaries. Elsewhere, \citet{kryscinski2020evaluating} used weak supervision---heuristic transformations used to intentionally introduce factual errors---to train a model to identify inaccuracies in outputs.

\citet{maynez2020faithfulness} 
enlisted humans to evaluate \emph{hallucinations}
(content found in a summary but not in its corresponding input) in automatically generated outputs. They report that for models trained on the XSUM dataset \cite{xsum}, over 70\% of summaries contain hallucinations. This corroborates other recent work \cite{falke2019ranking,wallace2020generating}, which has also found that ROUGE is a weak gauge of factuality. \citet{wang2020asking} proposed \emph{QAGS}, which uses automated question-answering to measure the consistency between reference and generated summaries. Elsewhere, \citet{xu2020fact} proposed evaluating textual factuality 
independent of surface realization via Semantic Role Labeling (SRL). Finally, \citet{pagnoni2021understanding} introduced the FRANK (meta-)benchmark for evaluating factuality metrics for summarization. While FRANK is tailored towards summarization-specific error categories including discourse, our ontology broadly reflects the goal of simplification (retaining content with simpler language) from the perspective of information precision, recall, and accuracy.

\begin{figure*}[t]
    \centering
    \includegraphics[width=\textwidth]{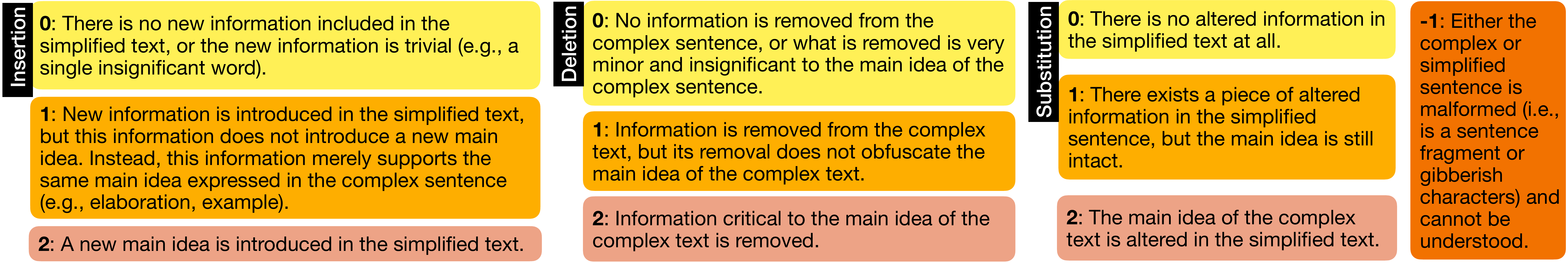}
    \caption{The full annotation scheme: 0: no/trivial change; 1: nontrivial but preserves main idea; 2: does not preserve main idea; -1: gibberish. The -1 label is applicable to all three categories.}
    \label{fig:annotation_scheme}
\end{figure*}

\section{Information Errors in Simplification}

Above we reviewed various recently proposed frameworks and methods for assessing the factual accuracy of automatically-generated \emph{summaries}. We aim in this work to similarly codify content errors in \emph{simplification}.

Below we describe broad categories of errors\footnote{We adapt a graded labeling scheme based on content and meaning preservation. For brevity, we use the word ``error'' as a generic term to refer to all the phenomena captured by our labeling scheme, even those that may be considered acceptable in some simplification systems.} we observed in simplification datasets and system outputs, and then use these to design annotation guidelines that formalize accuracy assessment (Section \ref{sec:crowdsourced_labeling}). Our analysis revealed three broad categories, illustrated in Table~\ref{tb:examples}:

\textbf{(1) Information Insertion:}
This occurs when information not mentioned in the complex sentence is inserted into---or \textit{hallucinated} in---its simplified counterpart. 
The insertion may be as small as mentioning a proper noun not in the complex sentence, or as large as introducing a new main idea. This category is similar to \textit{extrinsic hallucination} in the summarization literature~\cite{maynez2020faithfulness,goyal2021annotating}.

\textbf{(2) Information Deletion:}
This is when information in the complex sentence is omitted from the simplified sentence. 
A minor example of this is the reverse of the insertion case above, where an entity is mentioned by name in the complex sentence but only by pronoun in the simplified sentence.
    
\textbf{(3) Information Substitution:}
This is when information in the complex sentence is modified in the simplified sentence such that it changes the meaning. This category is broad, encompassing both alterations to the simplified sentence that directly contradict information in the complex sentence, and those that do not.

\begin{table}[t]
\centering
\small
\begin{tabular}{ll}
\toprule
\textbf{Category} & \textbf{Original/Simplified Sentences}              \\ \midrule
Insertion         & \textit{I went on a trip last week.}               \\
                  & \textit{I went on a trip to Alaska last week.}     \\
Deletion          & \textit{Yesterday I bought a bagel.}               \\
                  & \textit{I bought it.}                    \\
Substitution      & \textit{The shelter houses 100 cats and 200 dogs.} \\
                  & \textit{The shelter houses 200 cats and 200 dogs.} \\
                  \bottomrule
\end{tabular}
\vspace{-0.5em}
\caption{Illustrative examples of the three categories of information errors. Not from a real dataset.}
\label{tb:examples}
\end{table}

Because errors can co-occur, we adopt a multi-dimensional labeling scheme  that requires a different label to be provided for each category. Each category label specifies the severity of the error:
\textbf{0--no/trivial change; 1--nontrivial but preserves main idea; 2--doesn't preserve main idea; -1--gibberish,
specified in Figure~\ref{fig:annotation_scheme}.}
Table~\ref{tab:introexamples} shows level-2 examples from system outputs for insertion (examples 1-2), substitution (examples 3-4), and deletion (example 5).
Reference examples are discussed in Section~\ref{sec:Results_data}.

\vspace{-0.3em}
\paragraph{Interpretation as Precision and Recall}
In simplification one attempts to rewrite a given complex sentence to be simpler while preserving most of the information that it contains.
The categories above can be interpreted as errors in information precision (the fraction of content that also appears in the complex sentence) and recall (the fraction of content in the complex sentence preserved during simplification). With this interpretation, a ``false positive'' (affecting \emph{precision}) occurs when the simplified sentence contains information not present in the source, i.e., introduces a ``hallucination''. And a ``false negative'' (hindering \emph{recall}) is where the simplified sentence omits key information in the source.

\section{Data and Models}
\label{sec:datamodels}

We annotate data from the simplification datasets themselves (we will call these \emph{reference} examples), as well as from model-generated text. 
Thus we assess how the distribution of errors in the references compares to that of errors in system outputs and glean insights that might relate model architecture and training choices to the kinds of errors produced. 

\paragraph{Datasets.}
We annotated examples from the Wikilarge and Newsela~\cite{xu2015newsela, zhang2017sentence} datasets. 
These are commonly used in the literature, and so results have been reported on these corpora for a diverse collection of models. Wikilarge comprises 296K roughly-aligned sentences pairs from English Wikipedia and Simple English Wikipedia. Newsela~\cite{xu2015newsela} consists of 96K sentence pairs extracted from a dataset of news stories rewritten at 4 reading levels by professionals. To make analysis tractable in this work, we examine the simplest level for Newsela.

We annotated 400 pairs of (complex, simplified) sentences each from the validation and test sets for Newsela. For Wikilarge, we annotated 400 pairs from the validation set and 359 from the test set (this constitutes the entire test set).

\paragraph{Simplification Models.}
We annotated outputs generated by a collection of models on the same validation and test examples from Wikilarge and Newsela, respectively. We selected a set of models intended to be representative of different architectures and training methods.

More specifically, for RNN-based models we considered \texttt{Dress}~\cite{zhang2017sentence} and \texttt{EditNTS}~\cite{dong-etal-2019-editnts}. \texttt{Dress} is an LSTM model trained using  REINFORCE~\cite{williams1992simple} to minimize a reward function consisting of meaning preservation, simplicity, and fluency terms. \texttt{EditNTS} represents each sentence pair as a sequence of edit operations and directly learns these operations to perform simplification.

For Transformer-based architectures we evaluated two previously proposed models: \texttt{Access} \cite{access} and \texttt{ControlTS} \cite{controlts}. \texttt{Access} trains a randomly-initialized Transformer to generate simplifications parametrized by control tokens influencing traits like lexical complexity and length compression. \texttt{ControlTS} is a hybrid method that generates simplification candidates using grammatical rules and then applies a BERT-based~\cite{bert} paraphrasing model. In addition, we also fine-tuned \texttt{T5}~\cite{t5} for the simplification task, detailed in Appendix~\ref{app:t5}. \texttt{T5} is a Transformer-based model jointly pretrained both on unsupervised language modeling objectives and a host of supervised tasks including summarization and translation, all framed as text-to-text problems.

\section{Labeling with Mechanical Turk}
\label{sec:crowdsourced_labeling}
\paragraph{Annotation Procedure}
We use Amazon Mechanical Turk to acquire labels for reference examples from datasets, and for model-generated simplifications. To ensure that only annotators who understood our labeling scheme would be included, we released a qualification task consisting of 10 sentence pairs with perfect agreement among two of the authors, with detailed explanation of the labeling scheme, and required that annotators achieve at least 75\% accuracy on this set. 

After worker qualification, examples were released to only qualified workers, and each example was annotated by 3 workers. The final label for each category (insertion, deletion, substitution) was set to the majority label if one existed. If every annotator provided a different label for a given category, we removed this example for purposes of this category. For example, if annotators provided insertion labels of $\{1,1,2\}$ and deletion labels of $\{2,1,0\}$ for a specific instance, then this would not be assigned a deletion label, but would receive a ``final'' insertion label of 1. Workers were compensated \$10.00 per hour on the annotation task.

\noindent\textbf{Inter-annotator Agreement.}
We quantified the degree of inter-annotator agreement using 3 metrics, each capturing a different dimension of labeling consistency for each category: First, we report the percentage of examples that had a well-defined majority label for each category. Most annotators agreed on labels for the majority of examples (first column in Table~\ref{tb:majority_agreement}), meaning that very few annotations had to be discarded for any category.

\begin{table}[t]
\centering
\small
\begin{tabular}{l|lll}
\toprule
& \textbf{\% Majority} & \textbf{\% Majority}\\
\textbf{Category} & \textbf{Agreement} & \textbf{Agr.\ (non-zero)}\\
\midrule
Insertion         & 96 & 77\\
Deletion          & 96 & 92\\
Substitution      & 95 & 74\\
\bottomrule
\end{tabular}
\vspace{-0.5em}
\caption{Percentage of examples with majority annotator agreement for each category and percentage of examples with a majority nonzero label in which the majority of annotators agreed on the specific label.}
\label{tb:majority_agreement}
\vspace{-1em}
\end{table}

Because 0 was the most common label for all 3 categories, especially for the reference examples from the datasets, we also recorded the percentage of examples with \textit{majority non-zero annotations} that also have a well-defined majority label. 
For example, the labels $\{0,1,2\}$ are majority non-zero but do not correspond to a well-defined majority label, while $\{0,1,1\}$ satisfies both conditions. 
Table~\ref{tb:majority_agreement} (column 2) indicates that even among examples where most annotators agree that there is an error, the majority agree on a specific label of 1, 2, or -1. 

We also measured Krippendorff's alpha~\cite{krippendorffalpha} with an ordinal level of measurement (assigning the -1 label a value of 3 to indicate maximum severity). Dataset annotations for insertion enjoy moderate agreement ($\alpha=0.425$), those for deletion imply substantial agreement ($\alpha=0.639$), and those for substitution exhibit fair agreement ($\alpha=0.200$)~\cite{artstein2008inter}. The latter is possibly due to the clear majority label of 0 among substitution labels.

The \% majority agreement scores indicate that although the annotation scheme involves a degree of subjectivity in distinguishing between minor and major errors, with proper screening crowdsource workers can label text pairs with our annotation scheme consistently enough so that a well-defined label can be assigned to the vast majority of examples.

\section{Factuality of Reference Examples}
\label{sec:Results_data}

\begin{table}[t]
\centering
\small
\begin{tabular}{llllll}
\toprule
\textbf{Category} & \textbf{Dataset} & \textbf{0} & \textbf{1} & \textbf{2} & \textbf{-1} \\ \midrule
\rowcolor{light-gray}
Insertion         & Wikilarge        & 91.1       & 6.3        & 0.3        & 2.3         \\
                  & Newsela          & 68.2       & 20.2       & 11.1       & 0.5         \\
\rowcolor{light-gray}
Deletion          & Wikilarge        & 76.2       & 18.0       & 3.5        & 2.3         \\
                  & Newsela          & 15.8       & 40.8       & 42.9       & 0.5         \\
\rowcolor{light-gray}
Substitution      & Wikilarge        & 90.1       & 6.7        & 0.9        & 2.3         \\
                  & Newsela          & 94.9       & 3.8        & 0.8        & 0.5        \\
                  \bottomrule
\end{tabular}
\vspace{-0.5em}
\caption{Insertion, deletion, and substitution error distributions (\%) in Wikilarge and Newsela test datasets.}
\label{tb:dataset_dist}
\vspace{-1em}
\end{table}

\begin{table*}[t]
\centering
\small
\begin{tabular}{ll|lll|lll}
\toprule
& & \multicolumn{3}{c|}{\textbf{\% length change}} & \multicolumn{3}{c}{\textbf{Normalized edit distance}} \\
& & \textbf{Level 0}  & \textbf{Level 1}  & \textbf{Level 2} & \textbf{Level 0}  & \textbf{Level 1}  & \textbf{Level 2} \\ \midrule
\textbf{Insertion} &
Wikilarge & -5.0 (17.0) & 22.4 (36.9) & 7.1 (0.0) 
& 0.20 (0.20) & 0.55 (0.40) & 0.58 (0.0) \\
& Newsela & -39.4 (23.8) & -19.0 (36.9) & -38.3 (29.0) 
& 0.41 (0.17) & 0.51 (0.21) & 0.54 (0.04)\\
\midrule
\textbf{Deletion} &
Wikilarge & 2.8 (15.8) & -22.3 (18.9) & -35.9 (15.9) 
& 0.19 (0.23) & 0.35 (0.18) & 0.39 (0.14) \\
& Newsela & 1.5 (27.6) & -34.8 (23.1) & -49.6 (22.8) 
& 0.34 (0.31) & 0.46 (0.13) & 0.53 (0.10) \\
\bottomrule
\end{tabular}
\vspace{-0.5em}
\caption{\% length change (left) and normalized edit distances (right) in simplified sentences in each insertion and deletion error category (mean $\pm$ standard deviation).}
\label{tb:datasets_length_ed}
\vspace{-1em}
\end{table*}

\paragraph{Quantitative Analysis}
Table~\ref{tb:dataset_dist} reports distributions of acquired labels for information insertion, deletion, and substitution errors over the annotated reference examples.
Deletion errors are far more common than insertion errors in both datasets, though Wikilarge has fewer of both than Newsela. 
This is unsurprising, as one of the motivations for introducing the Newsela dataset was that it contains shorter and less syntactically-complex simplifications.
Reassuringly, there were very few substitution errors found in either dataset. 

Table~\ref{tb:datasets_length_ed} shows a clear positive correlation between length reduction and the severity of deletion errors present. As expected, sentences are shortened more substantially in Newsela than in Wikilarge. One the other hand, while Table~\ref{tb:datasets_length_ed} indicates that the examples with nonzero insertion labels collectively see a greater increase in length than those with no insertion errors, the mean length increase for level 2 examples is smaller than that for level 1.

Simplifications in Newsela are more abstractive~\cite{xu2015newsela}, i.e., simplified sentences copy fewer phrases verbatim from inputs. This can be quantified via normalized edit distance~\cite{Levenshtein1965}, which yielded a median of 0.46 for Newsela examples compared to the 0.38 for Wikilarge (after noise filtering described in Appendix~\ref{app:wikinoise}). Table~\ref{tb:datasets_length_ed} indicates that on average the more erroneous the insertion or deletion, the greater the normalized edit distance between the original and simplified sentences.

These results suggest that while reducing sentence length and rewording can be beneficial~\cite{klare1963measurement}, too much can negatively impact factuality.

\paragraph{Qualitative Analysis}
We also manually inspected insertion and deletion errors in both datasets, revealing clear patterns of deletion errors. Label 1 deletions by definition involve omissions of nonsalient details that do not much affect the meaning of the sentence, e.g.:
\advance\leftmargini -1em
\begin{quote}
\vspace{-0.25em}
\small
    \textbf{Original:} Mayfield wrote and sang on a string of message-oriented records, \emph{including ``Keep on Pushing" and ``People Get Ready."}\\
    \textbf{Simplified:} Mayfield wrote and sang on records that had a message. \hfill(\emph{Newsela, deletion-1})
\end{quote}
\vspace{-0.25em}

Label 2 deletions have two common manifestations across the datasets. The first involves deletion of the main clause and subsequent promotion of a secondary clause:
\vspace{-0.25em}
\begin{quote}
\small
    \textbf{Original:} ``Until you know how the sausage is made, you don't know how expensive it is to make that sausage,'' said Josh Updike, creative director of Rethink Leisure \& Entertainment, \emph{which is working on several projects in China and elsewhere in Asia.}\\ 
    \textbf{Simplified:} The company is working on several projects in China and Asia. \hfill(\emph{Newsela, deletion-2})
\end{quote}
\vspace{-0.25em}

Another common type of label 2 deletion involves removing a key (though often small) phrase that effectively reframes the entire sentence, e.g.:
\begin{quote}
\vspace{-0.25em}
\small
    \textbf{Original:} You may add a passage of up to five words as a Front-Cover Text, and a passage of up to 25 words as a Back-Cover Text, to the end of the list of Cover Texts \emph{in the Modified Version}. \\
    \textbf{Simplified:} You may add a passage of up to five words as a Front-Cover Text and a passage of up to 25 words as a Back-Cover Text to the end of the list of Cover Texts. \hfill(\emph{Wikilarge, deletion-2})
\end{quote}
\vspace{-0.25em}
By deleting \emph{in the Modified Version} (emphasis ours), the simplified sentence erroneously states that one may add front- and back-cover passages to the list of cover texts to the unmodified version, which is implicitly forbidden in the original.

Because of the small number of insertion errors on Wikilarge, we were unable to identify any meaningful trends. However, we observed patterns in Newsela for both levels 1 and 2 of insertions, pertaining to quotative phrases (e.g., inserting \emph{``experts said''} to the beginning of a sentence even though the original sentence did not mention an expert), and temporal phrases, e.g.:
\begin{quote}
\vspace{-0.25em}
\small
    \textbf{Original:} They could not afford to pay their son's roughly \$10,000 cost for classes at the University of Texas at Austin.\\
    \textbf{Simplified:} \emph{When he grew up,} they could not afford to pay \$10,000 for him to go to the University of Texas at Austin. \hfill(\emph{Newsela, insertion-1})
\end{quote}
\vspace{-0.25em}

Another error trend pertains to a change in specificity:
\vspace{-0.25em}
\begin{quote}
\small
    \textbf{Original:} Mutanabbi Street has always been \emph{a hotbed of dissent}.\\
    \textbf{Simplified:}  Mutanabbi Street has always been \emph{a place where protest marches are held}.
    \hfill(\emph{Newsela, insertion-2})
\end{quote}
\vspace{-0.5em}

We observed more contextually related errors for Newsela due to its style and its simplification process. Newsela documents were edited by professionals who rewrote the entire original document, and so information inserted or deleted could move from or to adjacent sentences. This preserves information for the whole document but causes problems at the sentence level. Also, compared to Wikilarge, Newsela's news articles naturally involve more complex discourse~\cite{van2013news}. These factors lead to relatively underspecified sentences~\cite{li2016improving} in the simplified text when they are taken out-of-context during training and evaluation. This observation calls for the inclusion of document context during simplification~\cite{sun2020helpfulness}, or performing decontextualization~\cite{choi2021decontextualization} before simplifying.

\begin{table*}[t]
\centering
\small
\begin{tabular}{ll|l|llll|llll|llll}
\toprule
& & & \multicolumn{4}{c|}{\textbf{Insertion}} & \multicolumn{4}{c|}{\textbf{Deletion}} & \multicolumn{4}{c}{\textbf{Substitution}} \\ \midrule
\textbf{Model} & \textbf{Dataset} & \textbf{SARI} & \textbf{0} & \textbf{1} & \textbf{2} & \textbf{-1} 
& \textbf{0} & \textbf{1} & \textbf{2} & \textbf{-1}
& \textbf{0} & \textbf{1} & \textbf{2} & \textbf{-1}\\ \midrule
\rowcolor{light-gray}
Dress          & Wikilarge  & 34.9     & 91.9       & 0.8        & 0.8        & 6.5         
                                  & 42.6       & 24.6       & 26.2       & 6.6         
                                  & 84.4       & 4.1        & 4.9        & 6.6         \\
\rowcolor{light-gray}
              & Newsela    & 34.5     & 90.5       & 0.0        & 0.0        & 9.5         
                                  & 29.9       & 29.2       & 32.1       & 9.7         
                                  & 67.4       & 6.5        & 15.9       & 10.1        \\ 
EditNTS        & Wikilarge  & 40.4     & 94.3       & 4.9        & 0.8        & 0.0         
                                  & 55.0       & 24.2       & 20.8       & 0.0         
                                  & 88.5       & 4.1        & 7.4        & 0.0         \\
              & Newsela    & 36.3     & 69.4       & 0.7        & 2.7        & 27.2        
                                  & 9.5        & 19.0       & 44.2       & 27.2        
                                  & 64.4       & 2.1        & 6.2        & 27.4        \\ 
\rowcolor{light-gray}
T5             & Wikilarge  & 34.9     & 96.8       & 1.6        & 0.8        & 0.8         
                                  & 81.6       & 14.4       & 3.2        & 0.8         
                                  & 97.6       & 1.6        & 0.0        & 0.8         \\
\rowcolor{light-gray}
              & Newsela    & 38.6     & 81.7       & 9.6        & 7.0        & 1.7         
                                  & 27.7       & 43.7       & 26.9       & 1.7         
                                  & 92.4       & 5.9        & 0.0        & 1.7         \\
Access         & Wikilarge  & 49.7     & 89.1       & 8.2        & 0.9        & 1.8         
                                  & 57.5       & 34.9       & 5.7        & 1.9         
                                  & 71.1       & 18.6       & 8.2        & 2.1         \\
\rowcolor{light-gray}
ControlTS      & Wikilarge  & 42.3     & 88.8       & 7.8        & 1.7        & 1.7        
                                  & 47.8       & 39.1       & 11.3       & 1.7        
                                  & 81.5       & 15.1       & 1.7        & 1.7        \\
\bottomrule
\end{tabular}
\caption{SARI and error distributions in system outputs manually evaluated.}
\label{tb:system_dist}
\end{table*}

\section{Factuality of System Outputs}

Table~\ref{tb:system_dist} shows the distributions of insertion, deletion, and substitution errors annotated in system outputs.\footnote{\texttt{DRESS} only released their Wikilarge outputs; \texttt{ControlTS} had different data splits for Newsela. We could not successfully reproduce their results for Newsela.} It also shows the standard simplification evaluation metric---SARI scores~\cite{sari}---for the annotated set. For the three models that reported both Wikilarge and Newsela outputs, the relative frequency of deletion errors between the two datasets appears to be preserved in model outputs, though for the RNN models errors are milder on Newsela and amplified on Wikilarge. 

A clear relationship between dataset and system output distributions does not exist for insertion and substitution errors. For \texttt{Dress} and \texttt{EditNTS}, this is due to the fact that the minor differences in insertion errors are dwarfed by the larger number of -1 (gibberish) labels assigned to Newsela outputs. Interestingly, outputs from the \texttt{T5} model were rarely labeled as -1 errors, so the difference in insertion errors is more apparent. In the case of substitution, the Newsela outputs for \texttt{Dress} and \texttt{T5} models show much higher rates of substitution errors than the Wikilarge outputs, despite the opposite being true for the datasets themselves. \texttt{EditNTS} does not show the same pattern, but again, the  high rate of -1 errors subsumes every other trend. One possible reason for this phenomenon could be that the higher abstractiveness of Newsela encourages models to rewrite the input sentence to a greater extent and destroy the original meaning in the process. In general the models produce substitution errors more frequently than are found in the dataset, meaning that they are introduced by the models themselves and not merely learned from the data.

\paragraph{Model comparisons}
There are a few differences in error distributions between the RNN-based and Transformer-based models, and between pre-trained vs.\ non-pretrained Transformer models. All three Transformer models have less severe deletion errors than the RNN models on Wikilarge, and in addition \texttt{T5} has lower deletion error rates on Newsela. Perhaps the most striking trend is that the Transformer models have far lower -1 gibberish errors than RNN-based models, even \texttt{Access}, which is not pre-trained on the language modeling task. \texttt{T5}---which has been pre-trained on large amounts of data---produced more insertion errors, while \texttt{Access} produced more substitution errors.

\paragraph{Quantitative Analysis}
We explore the relationships between the factuality annotations of system outputs and both length reduction and normalized edit distance. We briefly describe our findings here and defer numerical details to Appendix~\ref{app:system_tables}.

For every model except \texttt{Access}, there is a clear positive correlation between the severity of deletion errors and the degree of length reduction between the complex input and generated simplification. This is consistent with the trend observed for the datasets. No consistent relationships between length change and levels of insertion and substitution errors are exhibited by the system outputs. As in the case of length reduction, mean edit distances increase with the severity of deletion error with no consistent trends found for insertion and substitution labels.

\paragraph{Qualitative analysis} We also manually inspect model outputs, detailed in Appendix~\ref{app:modelanalysis}, and summarize main observations here.
As in the data, models also produce deletions ranging from single words and short phrases to clauses. For the two RNN models, \texttt{DRESS} and \texttt{EditNTS}, level 1 errors primarily consist of shorter deletion errors, which include pronoun errors and modifiers. 
Level 2 errors are almost always longer deletions, yet we did not observe the promotion of a subordinate clause to a main one as in the references, suggesting that models tend to follow syntactic rules more strictly. For \texttt{T5}, we additionally observe level 2 errors in which the model deletes a semantically critical word. We observed more error variability in the other two transformer models, \texttt{Access} and \texttt{ControlTS}.Models introduced varying numbers of insertion and substitution errors, but in inspection we did not observe any clear properties of these as a function of model type.

\section{Comparison with Existing Metrics}\label{sec:existing_metrics}
\paragraph{Relationship to SARI.}
SARI is the most popular metric used to evaluate text simplification models~\cite{sari}. For each model, we report Spearman's rank correlation coefficient~\cite{spearman} between SARI and each error category. As Table~\ref{tb:sari_corr} reports, there is only a weak correlation between SARI and the prevalence of information errors, and both the direction and magnitude of the correlation are highly dependent on model and dataset. This lack of correlation is unsurprising since SARI uses lexical overlap between the generated text with the reference text pair to judge simplification quality. This parallels the case with ROUGE in summarization~\cite{falke2019ranking,maynez2020faithfulness,wallace2020generating}.

\begin{table}[]
\centering
\small
\begin{tabular}{ll|LLL}
\toprule
\textbf{Model} & \textbf{Dataset} & \textbf{I} & \textbf{D} & \textbf{S} \\ \midrule
\rowcolor{light-gray}
Dress          & Wikilarge        & 0.038              & -0.041            & 0.156                 \\
\rowcolor{light-gray}
              & Newsela          & 0.105             & 0.267             & 0.258                 \\
EditNTS        & Wikilarge        & 0.011              & -0.275            & 0.034                 \\
              & Newsela          & -0.144              & -0.103             & -0.183                 \\
\rowcolor{light-gray}
T5             & Wikilarge        & -0.050              & 0.134             & 0.027                 \\
\rowcolor{light-gray}
              & Newsela          & -0.020             & -0.124            & 0.078                 \\
Access         & Wikilarge        & 0.035              & -0.026             & 0.057                 \\
\rowcolor{light-gray}
ControlTS      & Wikilarge        & 0.002              & -0.054            & 0.262                \\
\bottomrule
\end{tabular}
\vspace{-0.5em}
\caption{Spearman's rank correlation coefficients for SARI vs. each information error category ({\bf I}nsertion, {\bf D}eletion, {\bf S}ubstitution).}
\label{tb:sari_corr}
\vspace{-1em}
\end{table}

\paragraph{Measures of Semantic Similarity.}
Many existing text simplification systems attempt to address the problem of meaning preservation by using a semantic similarity score either directly in their loss/reward function or in a candidate ranking step~\cite{zhang2017sentence,kriz-etal-2019-complexity,Zhao_Chen_Chen_Yu_2020,controlts}. Additionally, some of these metrics have been included in recent factuality evaluation platforms in summarization~\cite{pagnoni2021understanding}. We explore the extent to which existing similarity methods detect information errors as outlined in our annotation scheme. We consider: (1) Jaccard similarity; (2) cosine similarity between averaged GloVe~\cite{glove} or ELMo~\cite{elmo} embeddings of the original and simplified sentences; (3) cosine similarity between Sentence-BERT~\cite{sentence-bert} embeddings; and (4) BERTScore~\cite{bertscore}.

As Table~\ref{tb:sem_similarity} indicates, the semantic similarity measures explored capture deletion errors quite well, while being a moderate indicator of insertion errors and a very weak one for substitution errors. Since deletion and substitution errors are common in most of the models we evaluated, the results indicate that better methods are needed to detect unacceptable deletions and intrinsic hallucinations in simplification outputs.

\begin{table}[]
\centering
\small
\begin{tabular}{l|LLL}
\toprule
{\textbf{Similarity Measure}} & \textbf{I} & \textbf{D} & \textbf{S} \\ \midrule
Jaccard Similarity                  & -0.385            & -0.695             & -0.101           \\
Cosine (GloVe)                  & -0.315            & -0.620             & -0.066           \\
Cosine (ELMo)                   & -0.325            & -0.582             & -0.065           \\
Cosine (Sentence BERT)                     & -0.375           & -0.724            & -0.182          \\
BERTScore                         & -0.400           & -0.748            & -0.125      \\
\bottomrule
\end{tabular}
\vspace{-0.5em}
\caption{Spearman's rank correlation coefficients for semantic similarity measures vs. each information error category ({\bf I}nsertion, {\bf D}eletion, {\bf S}ubstitution).}
\label{tb:sem_similarity}
\vspace{-1em}
\end{table}

\paragraph{Measures of Factuality.}
As in text simplification, the most common evaluation metrics used in text summarization like ROUGE do not adequately account for the factuality of model generations with respect to the input texts~\cite{kryscinski-etal-2019-neural}. For this reason, recent works have proposed model-based metrics to automatically assess factuality~\cite{falke-etal-2019-ranking,durmus-etal-2020-feqa,wang-qa-2020,factcc,goyal-dae}. We consider the following systems: (1) {\sc Fact-CC}, which is a BERT-based model trained on a synthetic dataset to classify text pairs as being factually inconsistent or not~\cite{factcc}, and (2) {\sc DAE}, which is another BERT-based model that classifies each dependency arc in the model output as entailing the source text or not~\cite{goyal-dae}. More specifically, for {\sc Fact-CC} we use the model's probability that each simplification example is inconsistent. For {\sc DAE} we use the average of the lowest $k$ probabilities that a dependency arc in the target sentence does not entail the source for $k=1,3,5$.

As Table~\ref{tb:fact_meas} indicates, both {\sc Fact-CC} and {\sc DAE}'s outputs correlate less with insertion and deletion annotations than even surface-level measures of semantic similarity like Jaccard similarity, though {\sc DAE} scores correlate better with substitution errors than do {\sc Fact-CC} and all evaluated measures of semantic similarity.

\begin{table}[]
\centering
\small
\begin{tabular}{l|lll}
\toprule
\textbf{Factuality Measure} & \textbf{I} & \textbf{D} & \textbf{S} \\ \hline
{\sc Fact-CC}                     & 0.311      & 0.418      & 0.165      \\
{\sc DAE}, $k=1$                  & 0.109      & 0.217      & 0.277      \\
{\sc DAE}, $k=3$                  & 0.110      & 0.213      & 0.271      \\
{\sc DAE}, $k=5$                  & 0.115      & 0.217      & 0.271      \\
\bottomrule
\end{tabular}
\caption{Spearman's rank correlation coefficients for factuality measures vs. each information error category ({\bf I}nsertion, {\bf D}eletion, {\bf S}ubstitution).}
\label{tb:fact_meas}
\end{table}

\section{Automatic Factuality Assessment}
Since manual annotation is costly and time-consuming, as a first step towards large-scale evaluation, we present an initial attempt at automating factuality assessment
by training a model on human annotations. To supplement training, we explore methods of generating synthetic data to improve model performance.

We framed automatic factuality assessment as a classification task in which a separate classifier is trained for each category (Insertion, Deletion, and Substitution), for each of the levels 0, 1, and 2. We treat the annotations used in our previous analyses as the test set and have additional data annotated to function as the training set for this task. We therefore collected a total of 1004 additional examples annotated across Wikilarge, Newsela, \texttt{Access} outputs on Wikilarge, and \texttt{T5} outputs on Newsela and Wikilarge. We fine-tuned RoBERTa~\cite{roberta} with a classification head. 

\paragraph{Synthetic Data Generation}
As Table~\ref{tab:training_label_counts} indicates, the validation dataset is both small and highly imbalanced, with very few level 2 insertion and substitution errors. To alleviate this issue, we experimented with a few methods of generating synthetic insertion and substitution errors on which to pretrain the model. We accomplished this by modifying each of the complex sentences in the validation set. To generate insertion errors, we replace names with pronouns and remove phrases from the source text to create target texts (information deletions) and then swap the source and target to produce information insertions. To generate substitutions, we change numbers in the source text, negate statements, and used BERT masking to perturb information in the sentence. We generated 10K examples in total; Appendix~\ref{sec:synthetic_data} describes these methods in greater detail.

\begin{table}[t]
    \centering
    \small
    \begin{tabular}{c|cc|cc|cc}
    \toprule
        & \multicolumn{2}{c|}{\textbf{Level 0}} & \multicolumn{2}{c|}{\textbf{Level 1}} & \multicolumn{2}{c}{\textbf{Level 2}}  \\
        \textbf{Category} & \textbf{\#} & \textbf{F1} & \textbf{\#} & \textbf{F1} & \textbf{\#} & \textbf{F1} \\
        \midrule
        Insertion & 823 & 87.9 & 104 & 36.6 & 40 & 30.4 \\
        Deletion & 413 & 84.2 & 356 & 57.1 & 204 & 52.1 \\
        Substitution & 810 & 82.7 & 110 & 19.8 & 33 & 9.5\\
    \bottomrule
    \end{tabular}
    \vspace{-0.5em}
    \caption{Annotated label counts in the training set, and F1 on the test set.}
    \label{tab:training_label_counts}
    \vspace{-1em}
\end{table}

\paragraph{Training and Evaluation}
The model is evaluated using the F1-scores with respect to each class (0,1,2), and when selecting checkpoints during training, the average of the label 1 and 2 F1
scores is used. The deletion model was trained directly on its training data, whereas the insertion and substitution models were initially pretrained on the synthetic datasets. Training details are provided in Appendix~\ref{app:roberta}.

\paragraph{Results}
Table~\ref{tab:training_label_counts} shows the test F1 scores achieved by the three classifiers. As expected, the deletion classifier achieved the best 1 and 2 F1 scores, likely due to the fact that the training dataset had plenty of level 1 and 2 deletion errors. Although the insertion and substitution datasets are similarly skewed, the insertion classifier significantly outperforms the substitution one. We found that using synthetic data is useful: without it, F1s for levels 1 and/or 2 are near 0 for insertion and substitution. Even with data augmentation, however, detecting errors is a challenging task.

\section{Conclusion}
We have presented an evaluation of the factuality of automated simplification corpora and model outputs, using an error typology with varied degrees of severity. We found that errors appear frequently in both references and generated outputs. In the datasets, deletion errors are quite frequent, with Newsela containing more than Wikilarge. The system outputs indicate that the models also tend to delete information, which is likely a behavior learned from the training data. Model outputs contain more substitution errors than the datasets, so that behavior is probably a model bias rather than something picked up from the data. 

Although we examined the two commonly used sentence-level datasets, factuality errors do extend to other domains and larger units of text. Our initial analysis of factuality in medical text simplification~\cite{devaraj2021paragraph} found errors of all three types, an indication that factual simplification is an open problem in such high-stake areas. The details of our analysis are in Appendix~\ref{app:cochrane}.

We also found that factuality errors are not well captured by existing metrics used in simplification such as SARI \cite{sari}. While semantic similarity metrics correlate with deletion errors, they poorly correlate with insertion or substitution. We further present an initial model for automatic factuality assessment, which we demonstrate is a challenging task.

\section*{Acknowledgements}

This work was partially supported by NSF grants IIS-1850153, IIS-2107524, IIS-1901117, as well as the National Institutes of Health (NIH), grant R01-LM012086.
We also acknowledge the Texas Advanced Computing Center (TACC) at UT Austin for providing the computational resources for many of the results within this paper. 
We are grateful to the anonymous reviewers for their comments and feedback.

\bibliography{main}
\bibliographystyle{acl_natbib}

\clearpage
\appendix

\section{Training details for the T5 simplification model}\label{app:t5}
We used the \texttt{T5} base architecture, which contains around 220M parameters. For both Newsela and Wikilarge, we trained the T5 model for 5 epochs with a batch size of 6 and constant learning rate of 3e-4. We prefixed each input text with the summarization prefix \texttt{summarize:}, since that was the task closest to simplification that the T5 model was pretrained on. Newsela simplifications were generated using nucleus sampling with $p=0.9$~\cite{holtzman2019curious}, and Wikilarge simplifications were generated using beam search with 6 beams.

\section{Noise filtering on Wikilarge}\label{app:wikinoise}
To filter out noisy alignments in the Wikilarge test set (when comparing the normalized edit distances between complex and simplified sentences in Newsela and Wikilarge), we employed the same method as used by ~\citet{xu2015newsela} to produce sentence-level alignments from the Newsela dataset, that is, we only keep sentence pairs if they have a Jaccard similarity of at least 0.4 if the simplification is one sentence long and 0.2 if it is longer than one sentence.

\section{Numerical Details for System Output Results}\label{app:system_tables}
Table~\ref{tb:system_length_change} shows the relationship between mean \% length reduction from input text to model output and the level of factuality errors present in the example. Table~\ref{tb:system_edit_distance} likewise shows the relationship between normalized edit distance between inputs and model outputs and factuality annotations.

\begin{table*}[t]
\centering
\small
\begin{tabular}{ll|lll|lll|lll}
\toprule
              &                  & \multicolumn{3}{c|}{\textbf{Insertion}}                         & \multicolumn{3}{c|}{\textbf{Deletion}} & \multicolumn{3}{c}{\textbf{Substitution}}         \\ \hline
\textbf{Model} & \textbf{Dataset} & \textbf{0} & \textbf{1}              & \textbf{2}               & \textbf{0}  & \textbf{1}  & \textbf{2} & \textbf{0} & \textbf{1} & \textbf{2}              \\ \hline
\rowcolor{light-gray}
Dress          & Wikilarge        & -20.7      & 6.3                     & -26.3                    & 0.11        & -26.8       & -47.4      & -21.0      & -10.5      & -15.1                   \\
\rowcolor{light-gray}
              & Newsela          & -29.4      & \multicolumn{1}{c}{---} & \multicolumn{1}{c|}{---} & -1.4        & -35.4       & -51.0      & -31.1      & -21.8      & -27.8                   \\
EditNTS        & Wikilarge        & -16.4      & 40.8                    & 72.7                     & 3.4         & -25.0       & -42.4      & -13.0      & -3.1       & -15.6                   \\
              & Newsela          & -41.6      & 33.3                    & -38.9                    & 0.8         & -39.4       & -51.9      & -40.2      & -57.9      & -32.7                   \\
\rowcolor{light-gray}
T5             & Wikilarge        & -4.1       & -4.6                    & -21.4                    & -0.04       & -22.2       & -30.5      & -4.5       & 0.0        & \multicolumn{1}{c}{---} \\
\rowcolor{light-gray}
              & Newsela          & -25.1      & -8.6                    & -25.4                    & 1.5         & -27.5       & -46.3      & -26.5      & 1.3        & \multicolumn{1}{c}{---} \\
Access         & Wikilarge        & -2.2       & 4.4                     & 0.0                      & 0.7         & -5.2        & 1.7        & -1.8       & -0.6       & -1.2                    \\
\rowcolor{light-gray}
ControlTS      & Wikilarge        & -10.6      & -5.9                    & -23.5                    & -1.5        & -16.2       & -28.2      & -11.1      & -5.5       & -22.3 \\    
\bottomrule
\end{tabular}
\caption{\% length change in system outputs (mean).}
\label{tb:system_length_change}
\end{table*}

\begin{table*}[t]
\centering
\small
\begin{tabular}{ll|lll|lll|lll}
\toprule
              &                  & \multicolumn{3}{c|}{\textbf{Insertion}}                         & \multicolumn{3}{c|}{\textbf{Deletion}} & \multicolumn{3}{c}{\textbf{Substitution}}         \\
\textbf{Model} & \textbf{Dataset} & \textbf{0} & \textbf{1}              & \textbf{2}               & \textbf{0}  & \textbf{1}  & \textbf{2} & \textbf{0} & \textbf{1} & \textbf{2}              \\ \hline
\rowcolor{light-gray}
Dress          & Wikilarge        & 0.23       & 0.06                    & 0.42                     & 0.03        & 0.32        & 0.49       & 0.23       & 0.22       & 0.17                    \\
\rowcolor{light-gray}
              & Newsela          & 0.29       & \multicolumn{1}{c}{---} & \multicolumn{1}{c|}{---} & 0.07        & 0.38        & 0.48       & 0.28       & 0.30       & 0.33                    \\
EditNTS        & Wikilarge        & 0.18       & 0.46                    & \multicolumn{1}{c|}{---} & 0.10        & 0.25        & 0.43       & 0.20       & 0.17       & 0.18                    \\
              & Newsela          & 0.36       & 0.33                    & \multicolumn{1}{c|}{---} & 0.10        & 0.37        & 0.46       & 0.37       & 0.42       & 0.25                    \\
\rowcolor{light-gray}
T5             & Wikilarge        & 0.08       & 0.53                    & \multicolumn{1}{c|}{---} & 0.04        & 0.30        & 0.56       & 0.09       & 0.09       & \multicolumn{1}{c}{---} \\
\rowcolor{light-gray}
              & Newsela          & 0.30       & 0.36                    & 0.56                     & 0.13        & 0.39        & 0.13       & 0.33       & 0.13       & \multicolumn{1}{c}{---} \\
Access         & Wikilarge        & 0.20       & 0.31                    & 0.14                     & 0.17        & 0.23        & 0.42       & 0.22       & 0.20       & 0.21                    \\
\rowcolor{light-gray}
ControlTS      & Wikilarge        & 0.24       & 0.43                    & 0.52                     & 0.12        & 0.38        & 0.50       & 0.27       & 0.24       & 0.52 \\
\bottomrule
\end{tabular}
\caption{Normalized edit distances in system outputs (mean).}
\label{tb:system_edit_distance}
\end{table*}

\section{Qualitative Analysis of System Outputs}\label{app:modelanalysis}
We also manually examined system outputs for error trends. Despite output variability for every model, two primary trends were observed in deletion errors across the models for both Wikilarge and, where available, Newsela. No trends could be drawn for insertion and substitution errors because of their infrequency.
The first type of deletion error, hence referred to as a “short”, is the deletion or change of a single word or short phrase, usually a modifier (such as an adjective, adverb, or serialized noun), but occasionally a noun, noun phrase, or verb. For example:
\begin{quote}
\small
    \textbf{Original:} The \emph{equilibrium} price for a certain type of labor is the wage. \\
	\textbf{Simplified:} The price of a certain type of labor is the wage.
	\hfill{\emph{(\texttt{ControlTS}, Wikilarge, deletion-1)}}
\end{quote}
When the word is changed rather than deleted, the replacing word is often less descriptive but can also be lateral. Shorts include pronoun errors, where a noun phrase is replaced with a pronoun. Note also that multiple, independent shorts may occur in an output and still receive a level 1 for deletion.
The second type of error, hence referred to as a “long”, is the deletion of a phrase, most commonly a prepositional phrase, or a \emph{subordinate} or \emph{coordinate} clause. For example:
\begin{quote}
\small
    \textbf{Original:} For Rowling, this scene is important because it shows Harry's bravery, \emph{and by retrieving Cedric's corpse, he demonstrates selflessness and compassion}.\\
	\textbf{Simplified:} For Rowling, this scene is important because it shows Harry's bravery.
	\hfill{\emph{(\texttt{Dress}, Wikilarge, deletion-2)}}
\end{quote}

Importantly, longs concerning clauses differ from the clause promotion error found in the datasets in that longs delete a \emph{subordinate} or \emph{coordinate} clause of the original while clause promotion errors delete the \emph{main} clause of the original. Multiple, independent longs rarely occur in one output; that is, if multiple secondary clauses are deleted, they are usually nested (likely because a sentence where this could happen would have a very complex structure, at least in English.)

\texttt{Access} and \texttt{ControlTS} had notable variability in the errors. Despite this, shorts were the most common error for label 1, with no notable presence of longs. These shorts were often not pronoun errors. Additionally, no trends could be noted for label 2 errors in these models. 
By contrast, nearly all of \texttt{Dress}’s errors fit into these two trends. Label 1 output errors primarily consisted of shorts, especially pronoun errors, though longs also occurred. Label 2 output errors were almost entirely longs. \texttt{EditNTS} and \texttt{T5} errors closely follow the trends found in Dress, though \texttt{T5} notably had several label 2 errors that were shorts, deleting a semantically-critical word. 

\section{Automatic Factuality Assessment}
Here we describe the details of generating synthetic data and training the three annotation classifiers.

\subsection{Synthetic Data Generation}
\label{sec:synthetic_data}
\noindent\textbf{Name Insertion.}
Each name of a person in the source text is replaced one at a time with a pronoun to create a target text. Then the source and target texts are swapped to simulate the insertion of a name in place of a pronoun. This text pair is labeled with a level 1 insertion.

\noindent\textbf{Phrase Insertion.}
Each phrase in the source text is deleted one at a time to create a shorter target text, and the source and target texts are swapped to simulate the insertion of a phrase. The insertion is labeled as a level 1 if the BERTScore of the texts is between 0.6 and 0.8, and it is labeled as 2 if it is between 0.2 and 0.4. If the score is not in either interval, the example is discarded. These thresholds were determined by manual inspection of the distribution of scores computed in Section~\ref{sec:existing_metrics}.

\noindent\textbf{Number Alteration.} We replace each number found in the source sentence one at a time with a random number of the same order of magnitude (e.g., $3\to 7$, $99\to 74$). This modification is labeled as a level 1 substitution.

\noindent\textbf{Statement Negation.}
Each auxiliary verb in the source text is negated one at a time to generate target texts. This modification is labeled as a level 1 substitution.

\noindent\textbf{BERT Masking.}
To generate level 1 substitutions, we randomly mask 2 tokens in the source text, pass the masked text through a BERT model, and fill the masked tokens with the third highest probability token in the output logits. To generate level 2 substitutions, we instead mask every fifth token in the source text and fill them with the fifth highest probability token indicated by the logits.

Once synthetic examples were generated, all the label 0 examples from the original training dataset were added. In the insertion synthetic dataset, level 1 labels significantly outnumbered level 2 labels, so only a random sample of them was included in the final dataset. Table~\ref{tab:synthetic_dataset_stats} shows the sizes and label distributions of the synthetic datasets. Some class imbalance was tolerated here since the number of examples for all levels was much larger than in the original training set and minority classes were oversampled during training.

\begin{table}[t]
    \centering
    \small
    \begin{tabular}{c|cccc}
    \toprule
        \textbf{Category} & \textbf{Level 0} & \textbf{Level 1} & \textbf{Level 2} & \textbf{Total} \\ \midrule
        Insertion & 823 & 1167 & 1167 & 3157\\
        Substitution & 810 & 4572 & 2008 & 7390 \\
    \bottomrule
    \end{tabular}
    \vspace{-0.5em}
    \caption{Sizes and label distributions of synthetic datasets.}
    \label{tab:synthetic_dataset_stats}
    \vspace{-1em}
\end{table}

\subsection{Model}
We fine-tune the pretrained base RoBERTa model architecture with a classification head. The model contains 12 hidden layers, a hidden size of 768, and 12 attention heads.

\subsection{Training Details}\label{app:roberta}
The insertion and substitution models were pretrained on an 80-20 train/dev split of their synthetic datasets for 10 epochs with a batch size of 64 and learning rate of 1e-4 and evaluated on the validation split every 100 steps. 

The best checkpoint was selected and then trained on an 80-20 split of its original dataset for 50 epochs with the same batch size and learning rate and evaluated every 10 steps. The best model from this round was finally fine-tuned on the entire training dataset for 1 epoch with the same batch size but a learning rate of 3e-5 before being evaluated on the test set. 

The deletion classifier was trained similarly, except that the pretraining step was omitted. 

In every stage of training, minority classes were oversampled in the training split until they matched the frequency of the most populous class.

\section{Case Study: Medical Texts}\label{app:cochrane}
We present an initial analysis of factuality in the context of medical text simplification \cite{devaraj2021paragraph}, a case where information accuracy is paramount. This task presents unique challenges given the complex, jargon-laden texts to be simplified. We evaluate a model proposed in recent work for medical text simplification \cite{devaraj2021paragraph}. This was trained by fine-tuning BART~\cite{lewis2020bart} on aligned \emph{paragraphs} of technical abstracts---plain English summaries from the Cochrane library, a database of systematic reviews of clinical trials. We annotated 10 randomly selected outputs from this model with respect to the original \emph{paragraphs}.\footnote{Note that while so far we have applied our annotation framework with respect to sentences, it is not tied to any specific linguistic unit.} Because the original texts are difficult to understand, we enlist a trained annotator (a senior in Linguistics and co-author of this work) to perform this evaluation.

Table~\ref{tab:cochrane} reports the number of error types observed across paragraphs. The error rate here is disconcerting: Out of the 10 paragraphs evaluated, we found 3 with at least one level-2 error, and 5 with more than one error. We provide examples or errors below. These findings further stresses the critical importance of factuality evaluation and improvement in simplification systems.

\begin{quote}
\small
    \textbf{Insertion}\\
    \textbf{Original:} [...] All studies were associated with methodological limitations. [...]\\
    \textbf{Simplified:} [...] All studies were \textcolor{red}{of poor quality and} had limitations in the way they were conducted. [...]
\end{quote}

\begin{quote}
\small
    \textbf{Substitution}\\
    \textbf{Original:} [...] There was \textcolor{blue}{no} difference in operating time or perioperative complication rates. [...] \\
    \textbf{Simplified:} [...] However, there was \textcolor{blue}{not enough evidence to determine if there was an important} difference in operative time or complication rates when compared to conventional surgery. [...]
\end{quote}

\begin{quote}
\small
    \textbf{Deletion}\\
    \textbf{Original:} Two trials with a total population of 1300 \textcolor{red}{amyotrophic lateral sclerosis} patients who were randomized to treatment with subcutaneous injections of recombinant human ciliary neurotrophic factor or placebo were examined in this review. [...] \\
    \textbf{Simplified:} In this review, we found two randomisedcontrolled trials with a total population of 1300 patients who were randomized to treatment with subcutaneous injections of recombinant human ciliary neurotrophic factor or placebo. [...]
\end{quote}

\begin{table}[t]
    \centering
    \small
    \begin{tabular}{c|ccc}
    \toprule
        \textbf{Category} & \textbf{Level 0} & \textbf{Level 1} & \textbf{Level 2}  \\ \midrule
        Insertion & 5 & 4 & 1 \\
        Deletion & 0 & 8 & 2 \\
        Substitution & 8 & 1 & 1\\
    \bottomrule
    \end{tabular}
    \vspace{-0.5em}
    \caption{Error judgments of the 10 example outputs from \citet{devaraj2021paragraph}.}
    \label{tab:cochrane}
    \vspace{-1em}
\end{table}

\end{document}